\documentclass[10pt,twocolumn,letterpaper]{article}

\usepackage{cvpr}           

\usepackage{graphicx}
\usepackage{amsmath}
\usepackage{amssymb}
\usepackage{booktabs}
\usepackage{multirow}
\usepackage{makecell} 
\usepackage{adjustbox}
\usepackage{tablefootnote}
\usepackage{xcolor} 
\usepackage{pifont}

\definecolor{myyellow}{HTML}{F8E924}
\definecolor{cvprblue}{RGB}{28,117,188}
\definecolor{aliceblue}{rgb}{0.94, 0.97, 1.0}
\definecolor{citegreen}{HTML}{009A55}
\definecolor{ggreen}{HTML}{3EC70B}
\newcommand{\cmark}{\textcolor{purple}{\ding{52}}}%
\newcommand{\xmark}{\textcolor{blue}{\ding{56}}}%

\usepackage[colorlinks=true,
            citecolor=citegreen,
            linkcolor=red,
            urlcolor=magenta, 
            pagebackref=true,
            breaklinks=true]{hyperref}

\usepackage[capitalize]{cleveref}
\usepackage{float}
\usepackage{blindtext}
\usepackage{graphicx}
\usepackage{caption}
\usepackage{bbding}

\title{Gaze-Guided Learning: Avoiding Shortcut Bias in Visual Classification}

\begin{document}

\title{Gaze-Guided Learning: Avoiding Shortcut Bias in Visual Classification}

\author{
  Jiahang Li$^{1}$ \quad Shibo Xue$^{1}$ \quad Yong Su$^{1}$\footnotemark[2]\\
  $^{1}$Tianjin Key Laboratory of Wireless Mobile Communications and Power Transmission,\\
  Tianjin Normal University \\
  {\footnotesize
  \texttt{lijiahang041119@gmail.com} \quad
  \texttt{18622594085@163.com} \quad
  \texttt{suyong@tju.edu.cn}
  }
}

\maketitle

\begin{abstract}
Inspired by human visual attention, deep neural networks have widely adopted attention mechanisms to learn locally discriminative attributes for challenging visual classification tasks. However, existing approaches primarily emphasize the representation of such features while neglecting their precise localization, which often leads to misclassification caused by shortcut biases. This limitation becomes even more pronounced when models are evaluated on transfer or out-of-distribution datasets. In contrast, humans are capable of leveraging prior object knowledge to quickly localize and compare fine-grained attributes, a capability that is especially crucial in complex and high-variance classification scenarios.
Motivated by this, we introduce Gaze-CIFAR-10, a human gaze time-series dataset, along with a dual-sequence gaze encoder that models the precise sequential localization of human attention on distinct local attributes. In parallel, a Vision Transformer (ViT) is employed to learn the sequential representation of image content. Through cross-modal fusion, our framework integrates human gaze priors with machine-derived visual sequences, effectively correcting inaccurate localization in image feature representations.
Extensive qualitative and quantitative experiments demonstrate that gaze-guided cognitive cues significantly enhance classification accuracy.
Both the dataset and code are publicly available at the \href{https://szyyjl.github.io/eye_tracking_data.github.io/}{project page} and on \href{https://github.com/rekkles2/Gaze-CIFAR-10}{this repository}, respectively.

\end{abstract}

\renewcommand{\thefootnote}{\fnsymbol{footnote}}
\footnotetext[2]{Corresponding author.}
\renewcommand{\thefootnote}{\arabic{footnote}}

%%%%%%%%% BODY TEXT

\section{Introduction}
\label{sec:intro}
\noindent
Computer Vision (CV), a core research area of artificial intelligence, aims to equip machines with the ability to \lq\lq see \rq\rq and interpret visual content, covering many tasks such as image classification \cite{1}, object detection \cite{2}, image segmentation \cite{3}, and video understanding \cite{4}. Recent advances in deep neural networks (DNNs), particularly with the introduction of attention mechanisms such as Transformers \cite{add37}, have significantly improved the modeling of global dependencies \cite{6}. Combined with large-scale datasets such as ImageNet \cite{7} and COCO \cite{8}, these models now surpass human performance on various vision benchmarks, driving substantial progress in the field.

\begin{figure}
    \centering
    \includegraphics[width=1.0\linewidth]{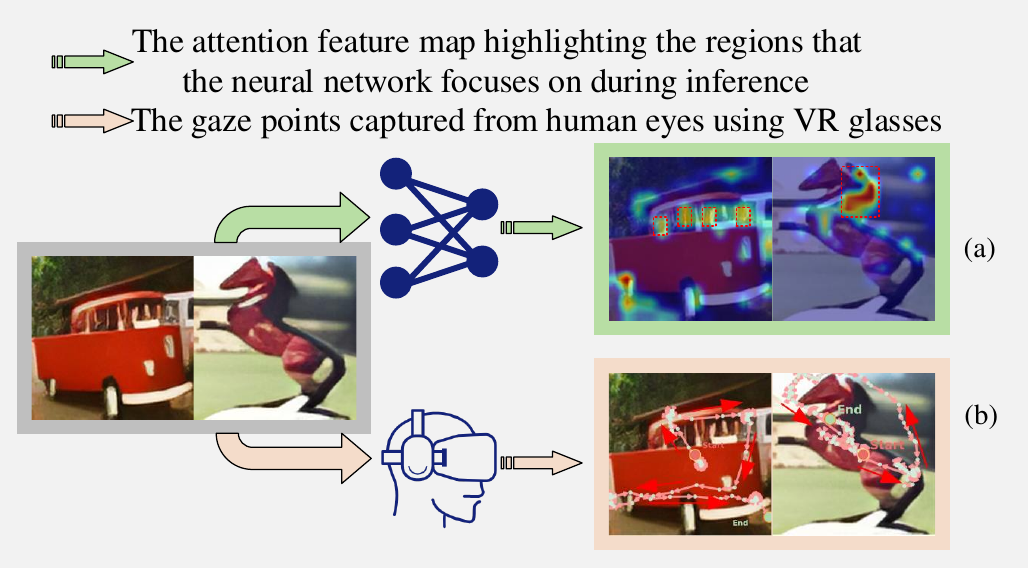}
    \caption{A toy example illustrating shortcut bias: (a) DNNs attention versus (b) human gaze under limited data scale and diversity.}
    \label{figca}
\end{figure}

However, the high-dimensionality, diversity, and irregular structure of visual data, along with the heterogeneity of vision tasks, present significant challenges to the development of Large-Scale Models (LSMs) similar to those in Natural Language Processing (NLP) \cite{add37}. Consequently, current research in CV remains largely task-specific, with models typically relying on customized training pipelines tailored to individual tasks. These models are highly dependent on the balance and richness of data-label pairs in the training set. 
In real-world scenarios, however, datasets are often imbalanced, with certain categories severely underrepresented \cite{add12}. For instance, in medical imaging \cite{9}, rare disease classes have significantly fewer samples than common ones, leading to degraded diagnostic performance on these minority categories. Similarly, in autonomous driving systems \cite{10}, models struggle to handle long-tail categories, such as uncommon traffic signs or infrequent road obstacles.
This data imbalance further exacerbates the problem of shortcut learning \cite{11}, where models fail to correctly localize the truly discriminative features of the target objects. Instead, they tend to exploit spurious correlations to minimize training loss, without learning visually consistent and semantically meaningful representations.

We present a toy example to further illustrate the bias introduced by shortcut learning. As shown in Figure~\ref{figca}, a ViT model pre-trained on ImageNet-21k is fine-tuned on the \textit{Gaze-CIFAR-10} dataset by updating only the classification head. During training, the model erroneously learns to associate the presence of humans with the labels \lq\lq bus\rq\rq and \lq\lq horse\rq\rq, due to biases in the visual patterns observed in the dataset. As a result, during inference, the model's attention focuses on misleading local features—such as people inside a bus or riders on horseback—as shown in Figure~\ref{figca}(a), ultimately leading to misclassification through the fine-tuned classification head.
In contrast, human cognition leverages prior knowledge to quickly attend to intrinsic and localized object features, such as texture and shape, rather than relying solely on data-driven statistical correlations. This enables robust recognition, even in ambiguous or previously unseen scenarios. As shown in Figure~\ref{figca}(b), human gaze gradually shifts toward the correct local discriminative regions, effectively overcoming the shortcut bias introduced during training. When the human-derived gaze guidance is integrated into the ViT's sequence representation, the misaligned token ordering is corrected, resulting in accurate classification.

To tackle these challenges, researchers have explored various solutions, including data augmentation, generative adversarial networks (GANs) \cite{12} for synthesizing minority class samples, self-supervised learning to exploit unlabeled data, and transfer learning to leverage knowledge from large-scale datasets \cite{13, add11}. While these methods have shown promise in mitigating class imbalance and data scarcity, they each have notable limitations. Data augmentation substantially increases computational overhead, and transfer learning often suffers from performance degradation due to domain shift.
More importantly, these approaches fail to address the core issue of shortcut bias, where the attention mechanisms in DNNs tend to focus on spurious or irrelevant local features rather than the truly discriminative ones. As a result, deep models still exhibit a significant gap in learning efficiency and robustness compared to human cognition.

Motivated by this, we construct a high-resolution variant of the widely used CIFAR-10 dataset, a standard benchmark for evaluating image classification performance. The improved resolution allows participants to engage in effortless visual recognition, enabling clearer observation of fine-grained details. Time-series gaze data collected during image viewing reveals the sequential nature of human cognitive processing and illustrates how visual attention progressively converges on locally discriminative features. 
To leverage gaze data for enhancing DNNs performance, we propose a gaze-guided baseline model for image recognition. This model integrates a dual-sequence gaze encoder that captures the sequential nature of human visual cognition in two dimensions: (1) the temporal progression of attention toward locally discriminative features, and (2) the spatial distribution of gaze points across the image. In parallel, a pre-trained ViT is employed to model the sequential representation of image content. Through cross-modal fusion, our framework aligns the visual information processing order guided by human gaze with the image-derived token sequence, thereby correcting misaligned sequential representations in the pre-trained model. Extensive qualitative and quantitative experiments demonstrate that incorporating gaze data enables DNNs to better align with human cognitive processes and significantly improves performance on image classification tasks.

\begin{figure*}[!t]
\begin{center}
\includegraphics[width=1\linewidth]{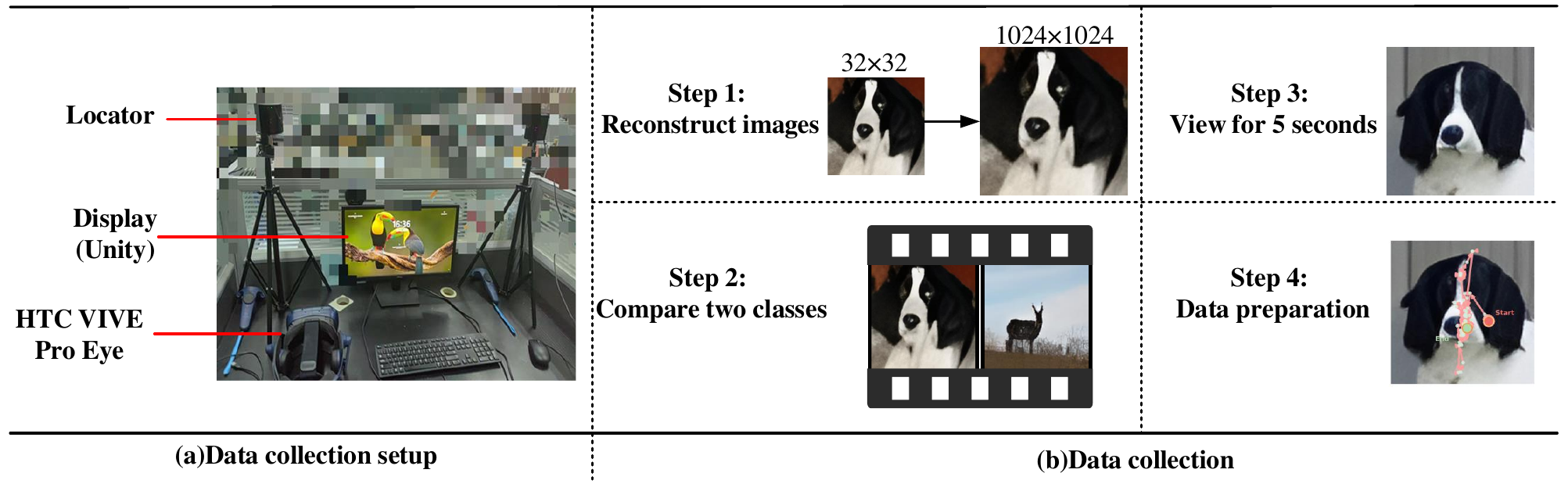}
\caption{Gaze data collection setup. (a) Overview of our data acquisition system. (b) Step 1: Reconstruct image resolution. Step 2: Participants freely view two randomly selected images from different categories. Step 3: One image is randomly re-sampled from the previously viewed categories and shown again for focused observation. Step 4: Gaze data is transmitted to the PC for processing.}
\label{figee}
\end{center}
\end{figure*}

\section{Related Work}
\noindent
Human gaze has emerged as a valuable modality for enhancing artificial intelligence systems across diverse domains such as NLP, CV, and Human-Robot Interaction (HRI). This section reviews recent advances in gaze-based AI, focusing on its applications in NLP, computer vision, and HRI, as well as comparative studies that integrate human and machine attention.

\subsection{Gaze in NLP}
\noindent
 Human gaze has been increasingly used to improve interpretability and performance in NLP tasks. Alaçam \textit{\textit{et al.}} \cite{17} introduced the GAZE4HATE dataset, combining hate speech annotations with gaze data to develop MEANION, a model integrating gaze features for the detection of hate speech. Their results demonstrated that gaze metrics such as dwell time significantly improve text-based model predictions by aligning more closely with human cognitive processes during annotation tasks. Similarly, Sood \textit{et al.} \cite{18} proposed a hybrid text saliency model, leveraging gaze-guided attention mechanisms to improve paraphrase generation and sentence compression tasks, achieving state-of-the-art results without task-specific gaze data. Eberle \textit{et al.} \cite{19} further analyzed the alignment between self-attention mechanisms in transformer models and human gaze during task-specific reading, finding that pre-trained models moderately correlate with human gaze, but underperform in capturing rare syntactic phenomena compared to cognitive models like the E-Z Reader.

\subsection{Gaze in CV}
\noindent
 Gaze data has been used to enhance interpretability and classification performance. Rong \textit{et al.} \cite{20} introduced the CUB-GHA dataset, which incorporates the human gaze for fine-grained image classification. They proposed Gaze Augmentation Training (GAT) and Knowledge Fusion Networks (KFN), showing significant performance improvements by integrating human attention into neural networks. Zhu \textit{et al.} \cite{21} advanced gaze-guided class activation mapping (GG-CAM) for chest X-ray classification, achieving higher interpretability and accuracy by aligning network attention with radiologists’ visual focus. Zhou \textit{et al.} \cite{22} extended gaze-based models to interaction recognition, introducing the Interactive-Gaze (IG) dataset and a zero-shot interaction prediction model, which outperformed traditional methods in understanding human-object interactions.

\subsection{Gaze in HRI}
\noindent
 Gaze-based intention recognition has shown potential for enhancing collaboration in HRI systems. Belardinelli \cite{23} provided a comprehensive survey of gaze-based methodologies for intention estimation, highlighting their utility in applications such as teleoperation and assistive robotics. Gaze was found to reliably predict user intentions, facilitating seamless human-robot coordination. The review emphasized the need to integrate cognitive principles of visuomotor control into technical systems to improve interaction design.

Comparison of human and machine attention patterns has provided valuable information on improving AI systems. Guo \textit{et al.} \cite{24} investigated the alignment between visual attention of humans and the saliency maps of reinforcement learning agents (RL) in Atari games. They identified discrepancies in attention patterns that contribute to performance gaps, highlighting the potential of human gaze data as a reference for training more interpretable and robust RL agents. Zhang \textit{et al.} \cite{25} provided a broader review of gaze-assisted AI, emphasizing the importance of gaze data in training attention mechanisms across domains, from vision and NLP to robotics.

\section{Data Processing and Collection}
\subsection{Collection Framework}
\noindent
To overcome the low resolution of the CIFAR-10 dataset ($32 \times 32$ pixels), which hinders reliable gaze data collection, we utilize Real-ESRGAN \cite{26}, a super-resolution model pre-trained on DIV2K, Flickr2K, and OutdoorSceneTraining datasets, to reconstruct the images to a resolution of $1024 \times 1024$ pixels. This enhanced resolution enables participants to perceive finer visual details, thereby supporting more accurate and consistent gaze tracking. 

An overview of our data collection setup is presented in Figure~\ref{figee}. Figure~\ref{figee}~(a) illustrates the gaze acquisition system, which consists of an HTC VIVE Pro Eye headset, a locator module, and a PC running Unity. A total of 20 participants were recruited to collect gaze data for all 60,000 images in the CIFAR-10 dataset.
As shown in \cite{add35}, when presented with two classes within a short time window, human observers tend to fixate on class-discriminative features. To capture such behavior while minimizing task-irrelevant exploratory gaze, each image was displayed for five seconds, with a two-second blank interval between images to prevent visual flicker and reduce eye strain. The gaze data captured by the HTC VIVE Pro Eye headset was transmitted to the PC and stored for subsequent analysis.
\begin{figure*}[t]
    \centering
    \includegraphics[width=1\linewidth]{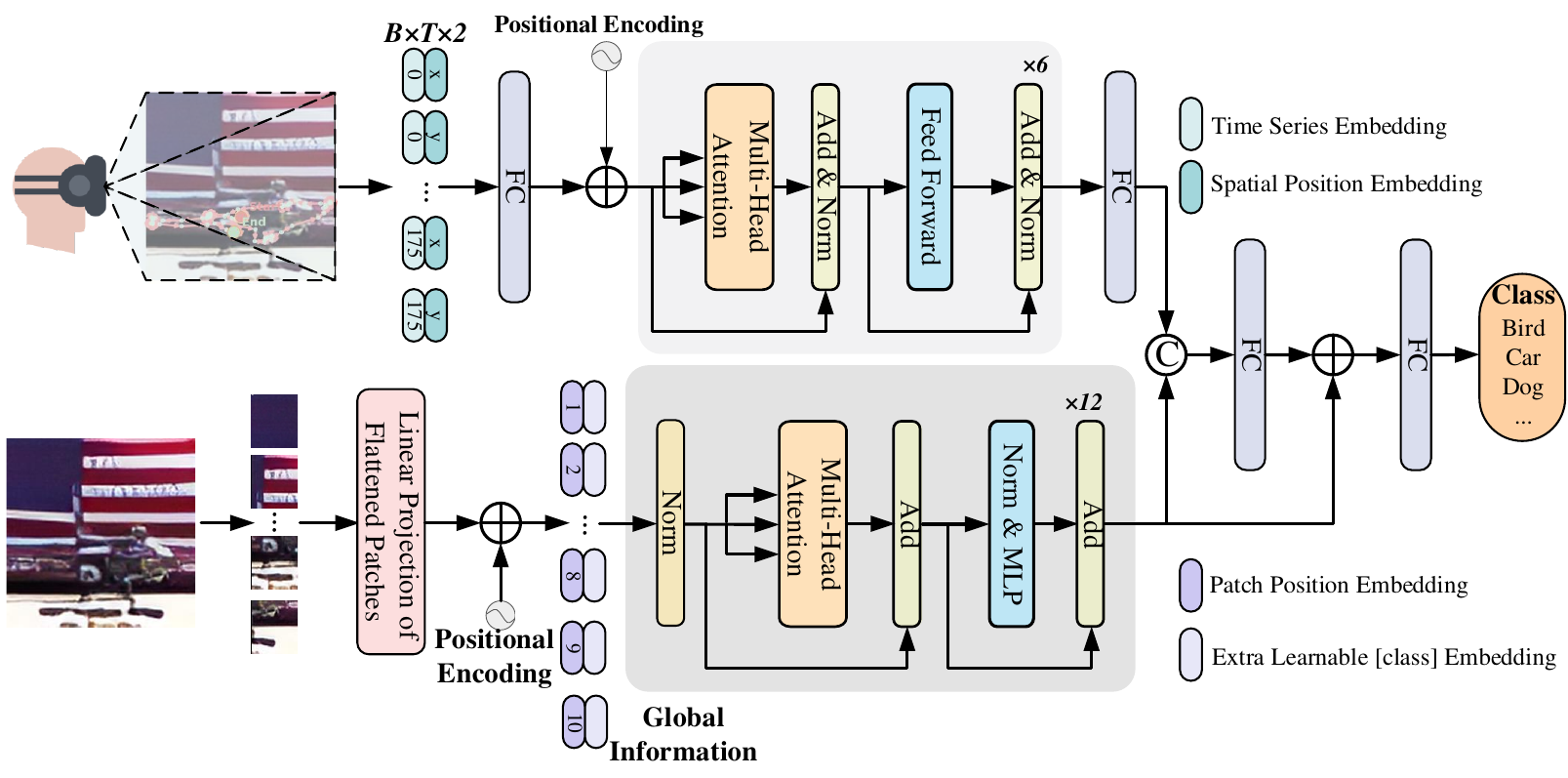}
    \caption{Gaze-guided cross-modal fusion network.}
    \label{model}
\end{figure*}

\subsection{Gaze-point preparation}
\noindent 
Unlike previous studies \cite{add36} that represent human gaze as Gaussian-distributed heatmaps, we model gaze as a sequential trajectory $\mathbf{G} \in \mathbb{R}^{176 \times 2}$. The second dimension, which corresponds to spatial coordinates, is normalized to the range $[0, 224]$ to match the input resolution of ViT. Finally, the dataset is randomly split into training and test sets with a ratio of $5:1$, resulting in 50,000 images for training and 10,000 for testing.

\section{Methodology}
\noindent
This section describes the proposed method, which leverages a dual-sequence gaze modeling mechanism to capture the sequential nature of human visual cognition along two dimensions. These gaze representations are then integrated with image token sequences extracted by a ViT backbone~\cite{27}. The overall pipeline comprises three main components: a dual-sequence gaze encoder, an image encoder, and a multi-modal feature fusion and classification module. An overview of the proposed model is shown in Figure~\ref{model}.

\subsection{Dual-Sequence Gaze Encoder}
\noindent
Given the gaze trajectory $\mathbf{G}$, represented as a matrix of size $176 \times 2$, where each row denotes both the temporal order of human attention to different local features and their corresponding spatial positions, we embed it into a high-dimensional feature space to capture the sequential characteristics of human visual cognition, as follows:

\textbf{Spatial Position Embedding. }The gaze matrix $\mathbf{G}$ is first passed through a fully connected (FC) layer to project it into a higher-dimensional representation: 
\begin{equation} 
\mathbf{\dot{x}}_{\text{1}} = \mathbf{W_1} \mathbf{G} + \mathbf{b_1}, 
\quad \mathbf{\dot{x}}_{\text{1}} \in \mathbb{R}^{176 \times 128} .
\end{equation}

\textbf{Dual-Sequence Feature Representation. }Human eye movement sequences are jointly driven by visual attention mechanisms and task demands, reflecting the information processing flow from global to local and from salient features to target regions. The temporal characteristics of gaze behavior reveal underlying cognitive strategies that ultimately guide attention toward discriminative local regions. These insights can inform the design of attention mechanisms and improve feature extraction efficiency in deep neural networks~\cite{itti1998model}. To integrate both the temporal dynamics of human cognition and the high-dimensional spatial representation of image content, we compute the Dual-Sequence correlation matrix as follows:
\begin{equation}
    \mathbf{A} = \text{Softmax} \left( \frac{\mathbf{Q} \mathbf{K}^\top}{\sqrt{d_h}} \right), \quad \mathbf{A} \in \mathbb{R}^{176 \times 176}.
\end{equation}

 We leverage the computed Dual-Sequence correlation matrix to capture the relative importance of gaze features across both temporal and spatial dimensions. This mechanism assigns adaptive weights to the sequential and positional components of human visual cognition, while effectively suppressing irrelevant or noisy signals. As a result, the model is guided to focus on critical gaze points and accurately localize the most informative regions in the image. The detailed process is described as follows:
\begin{equation}
    \text{head}_i = \mathbf{A} \cdot \mathbf{V}_i, \quad \text{head}_i \in \mathbb{R}^{176 \times d_h},
\end{equation}
\begin{equation}
\begin{aligned}
    \text{MHA}(\mathbf{X}) = \text{Concat}(\text{head}_1, \dots, \text{head}_H) W^O,\\
\end{aligned}
\end{equation}
\begin{equation}
\begin{aligned}
    \mathbf{\dot{x}_1'} = \text{LayerNorm}(\mathbf{\dot{x}_1} + \text{MHA}(\mathbf{\dot{x}_1})), 
    \quad \mathbf{\dot{x}_1'} \in \mathbb{R}^{176 \times d},
\end{aligned}
\end{equation}
\begin{equation}
\begin{aligned}
    \mathbf{\ddot{x}_2} = \text{LayerNorm}(\mathbf{\dot{x}_1'} + \text{ReLU}(\mathbf{\dot{x}_1'} \mathbf{W_1}' + \mathbf{b_1}') \mathbf{W_2}' + \mathbf{b_2}'),\\ 
    \quad \mathbf{\ddot{x}_2} \in \mathbb{R}^{176 \times d},
\end{aligned}
\end{equation}
where $d$ denotes the hidden dimension, and $d_h = d / H$ represents the dimensionality of each attention head. In this work, we set $H = 8$ and $d = 128$, and stack 6 identical Transformer encoder layers.

\textbf{Gazing Feature Aligning.} We transform $\mathbf{\ddot{x}_2}$ by aligning its spatial dimension with the output dimension of the ViT, and subsequently compress its temporal dimension to obtain a compact representation of sequential gaze information. This alignment ensures that the human gaze guidance is consistent with the image-derived token sequence, thereby correcting the erroneous sequential representations learned by the pre-trained model. Specifically, $\mathbf{\ddot{x}_2}$ is first passed through a fully connected (FC) layer to project it into a new feature space. Finally, we extract the feature at the first temporal position, resulting in a condensed vector representation $\mathbf{g} \in \mathbb{R}^{768}$.
\begin{equation}
    \mathbf{g'} = \mathbf{W}_2 \mathbf{\ddot{x}_2} + \mathbf{b}_2, \quad \mathbf{g'} \in \mathbb{R}^{176 \times 768},
\end{equation}

\begin{equation}
    \mathbf{g} = \mathbf{g'}[0, :], \quad \mathbf{g} \in \mathbb{R}^{768}.
\end{equation}

\subsection{Image Feature Extraction}
\noindent
The augmented image $\mathbf{I}$ is encoded using a standard ViT. The image is divided into patches, and each patch is linearly projected into a feature vector. The patch sequence is passed through multiple layers of Transformer blocks:
\begin{equation}
    \mathbf{\hat{I}} =\mathbf{ViT(I)}, \quad \mathbf{i} \in \mathbb{R}^{768}.
\end{equation}

\subsection{Multimodal Feature Fusion and Classification}
\noindent
To correct misaligned representations in the image token sequence using the spatiotemporal structure of human cognitive cues modeled by the Dual-Sequence Gaze Encoder, we apply a fusion mechanism combined with a residual connection. Specifically, we first concatenate the gaze feature $\mathbf{g}$ with the image feature $\mathbf{\hat{I}}$. Then, the fused representation is passed through a FC layer to integrate human cognitive guidance into the image sequence features. This enables accurate localization and enhancement of discriminative local regions based on the spatiotemporal structure of human attention, thereby improving classification performance. A skip connection is introduced by adding the original image feature $\mathbf{\hat{I}}$ back to the transformed representation, ensuring that global visual information is preserved.
\begin{equation}
    \mathbf{f'} = [\mathbf{g}; \mathbf{\hat{I}}], \quad \mathbf{f} \in \mathbb{R}^{1536},
\end{equation}
\begin{equation}
    \mathbf{f''} = (\mathbf{W}_3 \mathbf{f'} + \mathbf{b}_3) + \mathbf{\hat{I}}, \quad \mathbf{f}'' \in \mathbb{R}^{768},
\end{equation}
To predict class probabilities, the transformed features $\mathbf{f}''$ pass through a FC layer, followed by softmax activation. This process outputs the predicted probability distribution $\mathbf{\hat{y}}$ over $C$ classes:  
\begin{equation}
     \mathbf{\hat{y}} = \text{Softmax}(\mathbf{W}_4 \mathbf{f''} + \mathbf{b}_4), \quad \mathbf{\hat{y}} \in \mathbb{R}^C.
\end{equation}

\section{Experiment}
\noindent
We conducted both qualitative and quantitative experiments on the \textit{Gaze-CIFAR-10} dataset using multiple popular pre-trained backbones to evaluate the effectiveness of the proposed method. Standard metrics and visual results are reported to demonstrate performance improvements. In addition, we performed ablation studies and parameter sensitivity analyses to highlight the contributions of key components and to empirically validate our motivation: avoiding shortcut bias and guiding the model to attend to the correct locally discriminative features is essential for robust recognition.

\subsection{Experimental Setup}
\noindent
All experiments were conducted using an NVIDIA 3090 GPU with 24GB of memory, utilizing the PyTorch framework. The input images, originally at a resolution of $1024 \times 1024$, were resized to $224 \times 224$, followed by normalization with a mean and standard deviation of 0.5 for each RGB channel. The gaze trajectories, represented as $176 \times 2$ coordinate matrices, were linearly transformed into the range $[0,224]$, with sequences either padded or truncated to a fixed length of 176 points. For image feature extraction, we employed a Vision Transformer (ViT) with a patch size of $16 \times 16$, comprising six Transformer layers and a hidden dimension of 768.
The model training was conducted using Stochastic Gradient Descent (SGD)~\cite{28}, incorporating a momentum of $0.8$ and a weight decay of $5 \times 10^{-5}$. The learning rate was initially set to $0.001$ and updated following a cosine annealing schedule across 10 epochs. Specifically, the learning rate at epoch $x$, denoted as $\text{lr}(x)$, was computed as:
\begin{equation}
    \text{lr}(x) = \left[ \frac{1 + \cos\left(\frac{x \pi}{T}\right)}{2} \right] (1 - \eta_{\text{min}}) + \eta_{\text{min}},
\end{equation}
where $T = 10$ is the total number of epochs, and $\eta_{\text{min}} = 0.01$ represents the minimum learning rate threshold.
To enhance training stability and prevent overfitting, we employed a batch size of 32, along with dropout regularization at a rate of 0.1 within fully connected layers. Additionally, an early stopping mechanism with a patience threshold of 10 epochs was implemented to ensure optimal model generalization.

\begin{table}[h]
  \centering
  \small
  \caption{Impact of backbone and gaze encoder on accuracy (ACC$\uparrow$) for the Gaze-CIFAR-10 dataset. "W/O" refers to the ImageNet-21k pre-trained backbone fine-tuned on the proposed dataset, while other results correspond to backbones augmented with gaze features.}
  \label{table1}
  \begin{adjustbox}{max width=\textwidth}
  \resizebox{\linewidth}{!}{
    \begin{tabular}{lccccc}
      \toprule
      \textbf{Gaze Encoder} & \multicolumn{5}{c}{\textbf{Backbone}} \\ 
      \cmidrule{2-6}
      & \textbf{ViT} & \textbf{ResNet-50} & \textbf{MambaOut} & \textbf{RegNetY} & \textbf{ConvNeXtV2} \\
      \midrule
      \textbf{DSGE} & 84.20\% & 81.78\% & 84.01\% & 82.38\% & 85.90\% \\ 
      \textbf{MLP} & 81.95\% & 79.32\% & 83.62\% & 82.02\% & 85.67\% \\
      \textbf{W/O} & 81.18\% & 70.97\% & 83.28\% & 76.58\% & 79.46\% \\
      \bottomrule
    \end{tabular}}
    \end{adjustbox}
\end{table}

\subsection{Result on Gaze-CIFAR-10}
\noindent
\tablename~\ref{table1} investigates the impact of different gaze encoders on the classification accuracy of the \textit{Gaze-CIFAR-10} dataset across various backbone architectures. The three gaze encoding strategies compared are Dual-Sequence Gaze Encoder (DSGE), Multi-Layer Perceptron (MLP), and a baseline model without gaze features (denoted as "W/O"). Experimental results show that incorporating gaze features consistently improves model performance.
Specifically, DSGE achieves the highest accuracy of 85.90\% when combined with the ConvNeXtV2~\cite{29} backbone, outperforming other backbones such as ViT (84.20\%) and MambaOut~\cite{30} (84.01\%). Although the MLP encoder underperforms compared to DSGE, it still achieves a competitive maximum accuracy of 85.67\% with ConvNeXtV2. Its performance varies from 79.32\% (ResNet-50~\cite{31}) to 83.62\% (MambaOut). This suggests that MLP can only capture simple spatial mappings and fails to model the sequential nature of human visual learning, leading to inferior performance compared to DSGE, which effectively captures both temporal and spatial gaze dynamics.
In contrast, the baseline model without gaze encoding ("W/O") shows significantly lower accuracy, particularly with ResNet-50 (70.97\%) and RegNetY~\cite{32} (76.58\%), further highlighting the importance of gaze feature integration. Overall, these results validate that human gaze guidance can effectively enhance recognition performance across a variety of pre-trained backbone networks.

\subsection{Qualitative Results}
\begin{figure*}[h]
    \centering
    \includegraphics[width=1\linewidth]{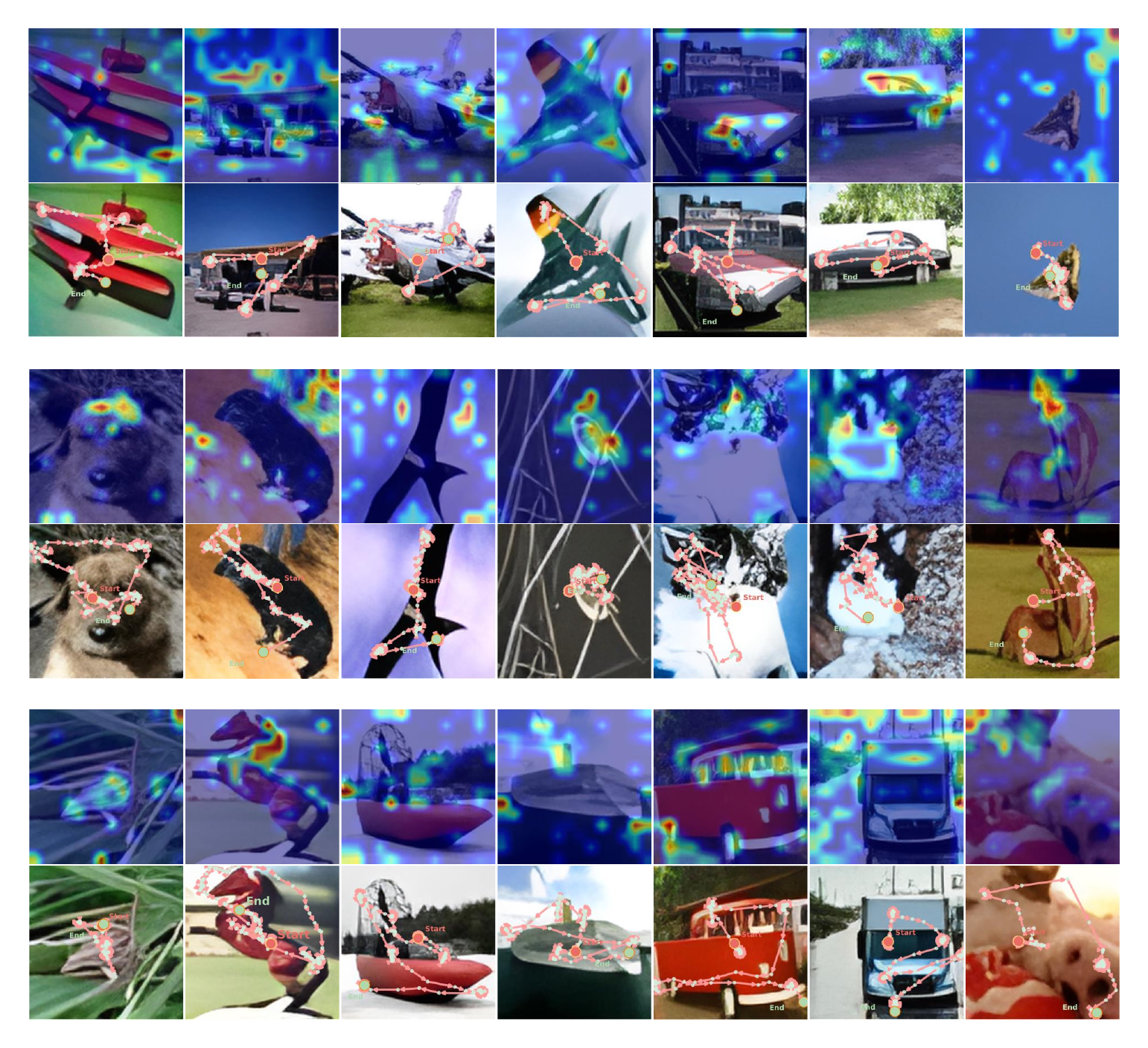}
    \caption{Comparison between ViT attention maps that lead to misclassification and human gaze points that guide correct classification. The red dot indicates the starting point of the gaze trajectory, while the green dot marks the end point.}
    \label{vit vs gaze}
\end{figure*}

\noindent
\figurename~\ref{vit vs gaze} presents three groups of visualizations. In each group, the top row displays the attention maps generated by the ViT model in cases of misclassification, while the bottom row shows the corresponding human gaze points. In contrast, our model correctly classifies the same samples by leveraging gaze guidance to attend to the appropriate discriminative regions.
Although the ViT model pre-trained on ImageNet-21k is capable of capturing local features effectively, it demonstrates limited generalization when fine-tuned on other datasets. Due to dataset bias, the model often fails to localize the truly discriminative regions, leading to incorrect predictions. In comparison, our model fuses human gaze information with the forward feature representations of the pre-trained ViT, effectively correcting misaligned token sequences and enhancing the generalization performance of the backbone on downstream tasks.

\begin{figure}
\begin{center}
\includegraphics[width=1\linewidth]{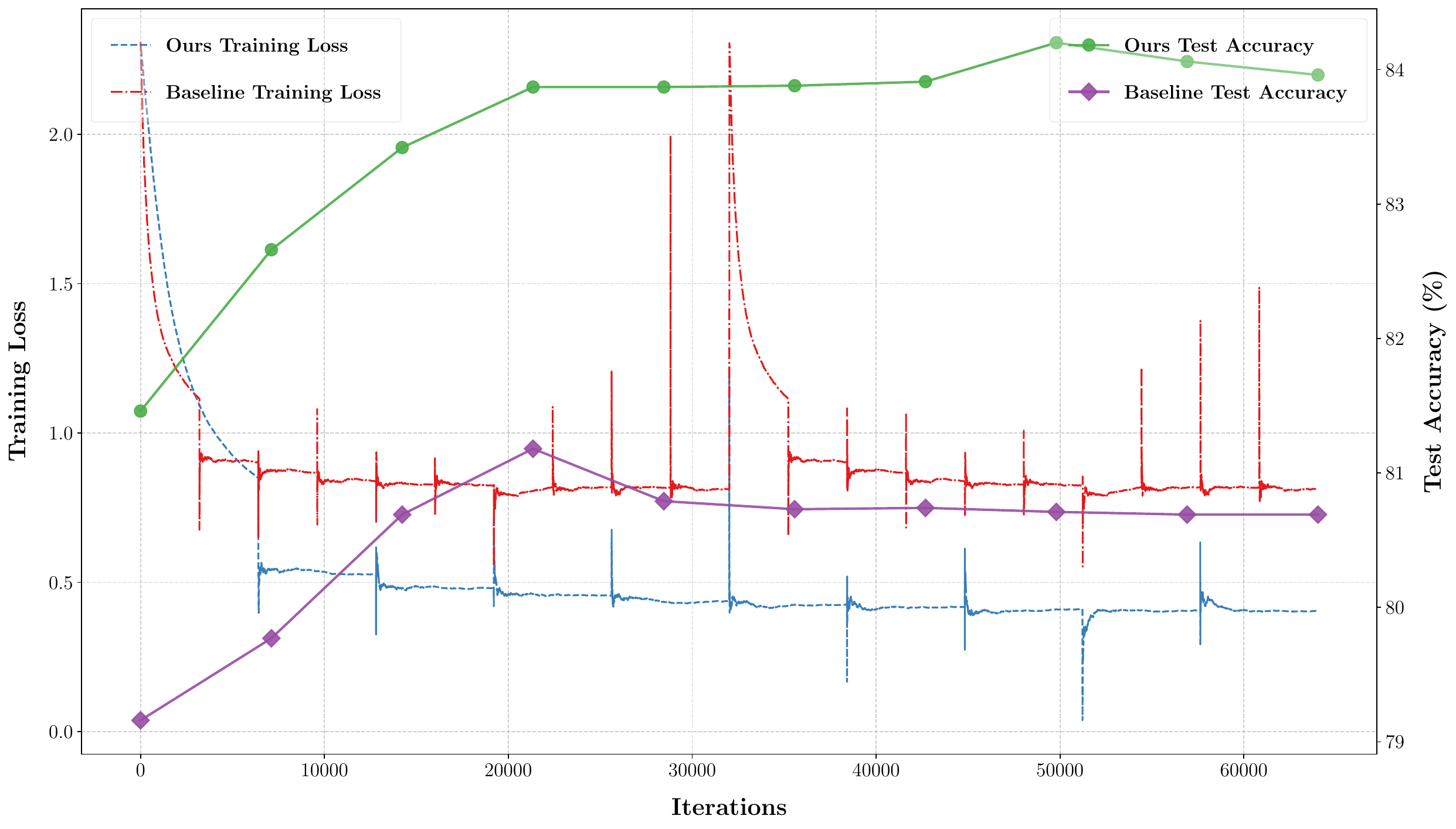}
\end{center}
   \caption{Training loss and test accuracy comparison between the proposed method and fine-tuned ViT.}
\label{fign}
\end{figure}

\figurename~\ref{fign} illustrates the training loss and test accuracy curves for the proposed method and the baseline model. The proposed method consistently shows a lower training loss throughout the training process, indicating better optimization and convergence stability compared to the baseline, which exhibits noticeable fluctuations. In terms of test accuracy, the proposed method achieves a peak accuracy of approximately 84\%, significantly outperforming the baseline, which plateaus around 81\%. These results highlight the effectiveness of the proposed enhancements, including the incorporation of gaze information and DSGE mechanisms, in improving both training stability and generalization performance.

\subsection{Ablation Study}
\noindent
\tablename~\ref{table3} presents the results of the ablation study, evaluating the contributions of three components: the Dual-Sequence Gaze Encoder (DSGE), Cross-Attention (CA), and the Fusion Layer. When using only the ViT backbone without gaze information, the model achieves an accuracy of 81.18\%. Introducing gaze trajectories through the DSGE module alone increases the accuracy to 83.29\%. However, adding the CA module while retaining DSGE leads to a slight drop in performance to 83.11\%. Incorporating all three components—DSGE, CA, and the Fusion Layer—yields an accuracy of 83.54\%. Notably, removing CA while preserving both DSGE and the Fusion Layer results in the best performance of 84.20\%.
Although cross-attention is commonly used in multimodal fusion, it proves suboptimal in our setting, where gaze signals are sparse and weakly aligned with image tokens. In our task, human gaze serves as a high-level supervisory cue, whereas ViT encodes dense patch-based visual representations. Applying cross-attention in this context may amplify modality mismatches and introduce feature entanglement, ultimately hindering performance.
In contrast, our fusion layer adopts a simple yet effective architecture that concatenates the gaze vector with the image token, followed by a fully connected layer and a skip connection. This design enables gaze-guided information injection without disrupting the structure of visual representations. It also promotes stable training and robust classification by preserving both global semantics and localized discriminative cues.

\begin{table}[h]
  \centering
  \small  
  \renewcommand{\arraystretch}{1.2}  
  \caption{Impact of the gaze feature's hidden dimension $h$ and Temporal Self-attention layer count $l$ on accuracy.}
  \label{table2}
  \begin{adjustbox}{max width=\linewidth} 
    \begin{tabular}{p{1.8cm} p{1.8cm} p{1.8cm} p{1.8cm}} 
      \toprule
      \multirow{2}{*}{\textbf{\textit{$h/l$}}} 
      & \multicolumn{3}{c}{\textbf{\textit{Layer Count $l$}}} \\ 
      \cmidrule(lr){2-4}
      & \textbf{\textit{$l=4$}} & \textbf{\textit{$l=6$}} & \textbf{\textit{$l=8$}} \\ 
      \midrule
      \textbf{$h=64$}  & 83.56 & 83.67 & 83.43 \\ 
      \textbf{$h=128$} & 83.32 & \bfseries 84.20 & 83.19 \\ 
      \textbf{$h=256$} & 83.68 & 83.84 & 83.77 \\ 
      \bottomrule
    \end{tabular}
  \end{adjustbox}
\end{table}

\begin{table}[h]
    \centering
    \small 
    \renewcommand{\arraystretch}{1.2}  
    \caption{Quantitative results of the ablation studies on our dataset. The symbol ``\cmark'' indicates the inclusion of the corresponding module, while ``\xmark'' denotes its exclusion.}
    \label{table3}
    \begin{adjustbox}{max width=\linewidth}  
        \begin{tabular}{p{1.8cm} p{1.8cm} p{2.0cm} p{0.7cm}} 
            \toprule
            \textbf{DSGE}  & \textbf{CA} & \textbf{Fusion Layer} & \textbf{ACC$\uparrow$}  \\ 
            \midrule
            \xmark & \xmark & \xmark & 81.18 \\ 
            \cmark & \xmark & \xmark & 83.29 \\ 
            \cmark & \cmark & \xmark & 83.11 \\  
            \cmark & \cmark & \cmark & 83.54 \\ 
            \cmark & \xmark & \cmark & \textbf{84.20} \\ 
            \bottomrule
        \end{tabular}
    \end{adjustbox}
\end{table}

\tablename~\ref{table2} illustrates the effect of varying the transformed feature dimension 
$h$ and the number of DSGE layers $l$ on the accuracy. Specifically, when \( h = 64 \), the accuracy increases slightly from 83.56\% to 83.67\% as \( l \) increases from 4 to 6, but drops to 83.43\% when \( l = 8 \). For \( h = 128 \), the highest accuracy of 84.20\% is achieved with \( l = 6 \), while the performance decreases to 83.19\% with \( l = 8 \). In contrast, when \( h = 256 \), the accuracy remains relatively stable, ranging from 83. 68\% to 83. 84\% at different values of \( l \). These findings highlight that the number of DSGE layers, especially \( l = 6 \), plays a crucial role in optimizing the performance of the model, especially when \( h = 128 \).

\section{Discussion \& Conclusion}
\noindent
In this work, we proposed a \textit{Gaze-CIFAR-10} dataset and a cross-modal gaze-image fusion method to mitigate shortcut learning issues in visual models. Our experiments demonstrate that incorporating gaze data, which capture human visual cognitive knowledge, effectively corrects the misrepresentation of local features, enhances performance, and improves model generalization. However, challenges remain in aligning human gaze information with visual features.
In future work, inspired by frameworks such as CLIP \cite{33} and ViLT \cite{add43}, we will explore multimodal alignment methods \cite{add44, add45, add46} and develop a gaze-vision alignment model that can be applied across multiple datasets, further enhancing the transferability of gaze information.
In addition, gaze information has significant application potential in small-scale datasets, particularly in few-shot learning \cite{add38, add39} and other domains that require expert annotation, such as medical imaging and other specialized fields. For example, in tasks such as medical image segmentation\cite{add40} and disease recognition \cite{add41, add42}, expert-provided gaze data can play a key role under conditions of data scarcity and high precision requirements. In the future, we will continue to explore gaze-based multimodal alignment strategies to further improve the robustness and generalization ability of visual models in these domains.

%%%%%%%%% REFERENCES
{\small
\bibliographystyle{ieee_fullname}
\bibliography{ref}

\begin{thebibliography}{10}\itemsep=-1pt

\bibitem{add46}
Hassan Akbari, Liangzhe Yuan, Rui Qian, Wei-Hong Chuang, Shih-Fu Chang, Yin Cui, and Boqing Gong.
\newblock Vatt: Transformers for multimodal self-supervised learning from raw video, audio and text.
\newblock {\em Advances in neural information processing systems}, 34:24206--24221, 2021.

\bibitem{17}
{\"O}zge Ala{\c{c}}am, Sanne Hoeken, and Sina Zarrie{\ss}.
\newblock Eyes don’t lie: Subjective hate annotation and detection with gaze.
\newblock In {\em Proceedings of the Conference on Empirical Methods in Natural Language Processing (EMNLP)}, pages 187--205, 2024.

\bibitem{add36}
Xiao Bai, Pengcheng Zhang, Xiaohan Yu, Jin Zheng, Edwin~R Hancock, Jun Zhou, and Lin Gu.
\newblock Learning from human attention for attribute-assisted visual recognition.
\newblock {\em IEEE Transactions on Pattern Analysis and Machine Intelligence}, 2024.

\bibitem{23}
Anna Belardinelli.
\newblock Gaze-based intention estimation: Principles, methodologies, and applications in hri.
\newblock {\em ACM Transactions on Human-Robot Interaction}, 13(3):1--30, 2024.

\bibitem{13}
Alexander Brown, Nenad Tomasev, Jan Freyberg, Yuan Liu, Alan Karthikesalingam, and Jessica Schrouff.
\newblock Detecting shortcut learning for fair medical ai using shortcut testing.
\newblock {\em Nature Communications}, 14(1):4314, 2023.

\bibitem{add44}
Qi Chen, Mingkui Tan, Yuankai Qi, Jiaqiu Zhou, Yuanqing Li, and Qi Wu.
\newblock V2c: Visual voice cloning.
\newblock In {\em Proceedings of the IEEE/CVF Conference on Computer Vision and Pattern Recognition}, pages 21242--21251, 2022.

\bibitem{7}
Jia Deng, Wei Dong, Richard Socher, Li-Jia Li, Kai Li, and Li Fei-Fei.
\newblock Imagenet: A large-scale hierarchical image database.
\newblock In {\em Proceedings of the IEEE Conference on Computer Vision and Pattern Recognition (CVPR)}, pages 248--255, 2009.

\bibitem{9}
Tribikram Dhar, Nilanjan Dey, Surekha Borra, and R~Simon Sherratt.
\newblock Challenges of deep learning in medical image analysis—improving explainability and trust.
\newblock {\em IEEE Transactions on Technology and Society}, 4(1):68--75, 2023.

\bibitem{27}
Alexey Dosovitskiy.
\newblock An image is worth 16x16 words: Transformers for image recognition at scale.
\newblock {\em arXiv preprint arXiv:2010.11929}, 2020.

\bibitem{19}
Oliver Eberle, Stephanie Brandl, Jonas Pilot, and Anders S{\o}gaard.
\newblock Do transformer models show similar attention patterns to task-specific human gaze?
\newblock In {\em Proceedings of the Annual Meeting of the Association for Computational Linguistics (ACL)}, pages 4295--4309, 2022.

\bibitem{11}
Robert Geirhos, J{\"o}rn-Henrik Jacobsen, Claudio Michaelis, Richard Zemel, Wieland Brendel, Matthias Bethge, and Felix~A Wichmann.
\newblock Shortcut learning in deep neural networks.
\newblock {\em Nature Machine Intelligence}, 2(11):665--673, 2020.

\bibitem{24}
Suna~Sihang Guo, Ruohan Zhang, Bo Liu, Yifeng Zhu, Dana Ballard, Mary Hayhoe, and Peter Stone.
\newblock Machine versus human attention in deep reinforcement learning tasks.
\newblock {\em Advances in Neural Information Processing Systems (NeurIPS)}, 34:25370--25385, 2021.

\bibitem{31}
Kaiming He, Xiangyu Zhang, Shaoqing Ren, and Jian Sun.
\newblock Deep residual learning for image recognition.
\newblock In {\em Proceedings of the IEEE Conference on Computer Vision and Pattern Recognition (CVPR)}, pages 770--778, 2016.

\bibitem{itti1998model}
Laurent Itti, Christof Koch, and Ernst Niebur.
\newblock A model of saliency-based visual attention for rapid scene analysis.
\newblock {\em IEEE Transactions on Pattern Analysis and Machine Intelligence}, 20(11):1254--1259, 1998.

\bibitem{add35}
Nour Karessli, Zeynep Akata, Bernt Schiele, and Andreas Bulling.
\newblock Gaze embeddings for zero-shot image classification.
\newblock In {\em Proceedings of the IEEE conference on computer vision and pattern recognition}, pages 4525--4534, 2017.

\bibitem{add42}
Daniel~S Kermany, Michael Goldbaum, Wenjia Cai, Carolina~CS Valentim, Huiying Liang, Sally~L Baxter, Alex McKeown, Ge Yang, Xiaokang Wu, Fangbing Yan, et~al.
\newblock Identifying medical diagnoses and treatable diseases by image-based deep learning.
\newblock {\em cell}, 172(5):1122--1131, 2018.

\bibitem{add43}
Wonjae Kim, Bokyung Son, and Ildoo Kim.
\newblock Vilt: Vision-and-language transformer without convolution or region supervision.
\newblock In {\em International conference on machine learning}, pages 5583--5594. PMLR, 2021.

\bibitem{4}
Kunchang Li, Xinhao Li, Yi Wang, Yinan He, Yali Wang, Limin Wang, and Yu Qiao.
\newblock Videomamba: State space model for efficient video understanding.
\newblock In {\em Proceedings of the European Conference on Computer Vision (ECCV)}, pages 237--255, 2025.

\bibitem{8}
Tsung-Yi Lin, Michael Maire, Serge Belongie, James Hays, Pietro Perona, Deva Ramanan, Piotr Doll{\'a}r, and C~Lawrence Zitnick.
\newblock Microsoft coco: Common objects in context.
\newblock In {\em Proceedings of the European Conference on Computer Vision (ECCV)}, pages 740--755, 2014.

\bibitem{3}
Shervin Minaee, Yuri Boykov, Fatih Porikli, Antonio Plaza, Nasser Kehtarnavaz, and Demetri Terzopoulos.
\newblock Image segmentation using deep learning: A survey.
\newblock {\em IEEE Transactions on Pattern Analysis and Machine Intelligence}, 44(7):3523--3542, 2021.

\bibitem{10}
Aditya Prakash, Kashyap Chitta, and Andreas Geiger.
\newblock Multi-modal fusion transformer for end-to-end autonomous driving.
\newblock In {\em Proceedings of the IEEE/CVF Conference on Computer Vision and Pattern Recognition (CVPR)}, pages 7077--7087, 2021.

\bibitem{33}
Alec Radford, Jong~Wook Kim, Chris Hallacy, Aditya Ramesh, Gabriel Goh, Sandhini Agarwal, Girish Sastry, Amanda Askell, Pamela Mishkin, Jack Clark, et~al.
\newblock Learning transferable visual models from natural language supervision.
\newblock In {\em Proceedings of the International Conference on Machine Learning (ICML)}, pages 8748--8763, 2021.

\bibitem{32}
Ilija Radosavovic, Raj~Prateek Kosaraju, Ross Girshick, Kaiming He, and Piotr Doll{\'a}r.
\newblock Designing network design spaces.
\newblock In {\em Proceedings of the IEEE/CVF Conference on Computer Vision and Pattern Recognition (CVPR)}, pages 10428--10436, 2020.

\bibitem{28}
Herbert Robbins and Sutton Monro.
\newblock A stochastic approximation method.
\newblock {\em The annals of mathematical statistics}, pages 400--407, 1951.

\bibitem{20}
Yao Rong, Wenjia Xu, Zeynep Akata, and Enkelejda Kasneci.
\newblock Human attention in fine-grained classification.
\newblock {\em arXiv preprint arXiv:2111.01628}, 2021.

\bibitem{add38}
Yisheng Song, Ting Wang, Puyu Cai, Subrota~K Mondal, and Jyoti~Prakash Sahoo.
\newblock A comprehensive survey of few-shot learning: Evolution, applications, challenges, and opportunities.
\newblock {\em ACM Computing Surveys}, 55(13s):1--40, 2023.

\bibitem{18}
Ekta Sood, Simon Tannert, Philipp M{\"u}ller, and Andreas Bulling.
\newblock Improving natural language processing tasks with human gaze-guided neural attention.
\newblock {\em Advances in Neural Information Processing Systems (NeurIPS)}, 33:6327--6341, 2020.

\bibitem{add12}
Yong Su, Yuyu Tan, Simin An, and Meng Xing.
\newblock Anomalies cannot materialize or vanish out of thin air: A hierarchical multiple instance learning with position-scale awareness for video anomaly detection.
\newblock {\em Expert Systems with Applications}, 254:124392, 2024.

\bibitem{add11}
Yong Su, Yuyu Tan, Meng Xing, and Simin An.
\newblock Vpe-wsvad:visual prompt exemplars for weakly-supervised video anomaly detection.
\newblock {\em Knowledge-Based Systems}, 299:111978, 2024.

\bibitem{add41}
Edna~Chebet Too, Li Yujian, Sam Njuki, and Liu Yingchun.
\newblock A comparative study of fine-tuning deep learning models for plant disease identification.
\newblock {\em Computers and Electronics in Agriculture}, 161:272--279, 2019.

\bibitem{add37}
Ashish Vaswani, Noam Shazeer, Niki Parmar, Jakob Uszkoreit, Llion Jones, Aidan~N Gomez, {\L}ukasz Kaiser, and Illia Polosukhin.
\newblock Attention is all you need.
\newblock {\em Advances in neural information processing systems}, 30, 2017.

\bibitem{1}
Fei Wang, Mengqing Jiang, Chen Qian, Shuo Yang, Cheng Li, Honggang Zhang, Xiaogang Wang, and Xiaoou Tang.
\newblock Residual attention network for image classification.
\newblock In {\em Proceedings of the IEEE/CVF Conference on Computer Vision and Pattern Recognition (CVPR)}, pages 3156--3164, 2017.

\bibitem{add40}
Risheng Wang, Tao Lei, Ruixia Cui, Bingtao Zhang, Hongying Meng, and Asoke~K Nandi.
\newblock Medical image segmentation using deep learning: A survey.
\newblock {\em IET image processing}, 16(5):1243--1267, 2022.

\bibitem{26}
Xintao Wang, Liangbin Xie, Chao Dong, and Ying Shan.
\newblock Real-esrgan: Training real-world blind super-resolution with pure synthetic data.
\newblock In {\em Proceedings of the IEEE/CVF International Conference on Computer Vision (ICCV)}, pages 1905--1914, 2021.

\bibitem{add39}
Yaqing Wang, Quanming Yao, James~T Kwok, and Lionel~M Ni.
\newblock Generalizing from a few examples: A survey on few-shot learning.
\newblock {\em ACM computing surveys (csur)}, 53(3):1--34, 2020.

\bibitem{29}
Sanghyun Woo, Shoubhik Debnath, Ronghang Hu, Xinlei Chen, Zhuang Liu, In~So Kweon, and Saining Xie.
\newblock Convnext v2: Co-designing and scaling convnets with masked autoencoders.
\newblock In {\em Proceedings of the IEEE/CVF Conference on Computer Vision and Pattern Recognition (CVPR)}, pages 16133--16142, 2023.

\bibitem{12}
Wanqian Yang, Polina Kirichenko, Micah Goldblum, and Andrew~G Wilson.
\newblock Chroma-vae: Mitigating shortcut learning with generative classifiers.
\newblock {\em Advances in Neural Information Processing Systems (NeurIPS)}, 35:20351--20365, 2022.

\bibitem{30}
Weihao Yu and Xinchao Wang.
\newblock Mambaout: Do we really need mamba for vision?
\newblock {\em arXiv preprint arXiv:2405.07992}, 2024.

\bibitem{add45}
Hang Zhang, Xin Li, and Lidong Bing.
\newblock Video-llama: An instruction-tuned audio-visual language model for video understanding.
\newblock {\em arXiv preprint arXiv:2306.02858}, 2023.

\bibitem{25}
Ruohan Zhang, Akanksha Saran, Bo Liu, Yifeng Zhu, Sihang Guo, Scott Niekum, Dana Ballard, and Mary Hayhoe.
\newblock Human gaze assisted artificial intelligence: A review.
\newblock In {\em Proceedings of the International Joint Conference on Artificial Intelligence (IJCAI)}, volume 2020, page 4951, 2020.

\bibitem{6}
Yian Zhao, Wenyu Lv, Shangliang Xu, Jinman Wei, Guanzhong Wang, Qingqing Dang, Yi Liu, and Jie Chen.
\newblock Detrs beat yolos on real-time object detection.
\newblock In {\em Proceedings of the IEEE/CVF Conference on Computer Vision and Pattern Recognition (CVPR)}, pages 16965--16974, 2024.

\bibitem{22}
Yuchen Zhou, Linkai Liu, and Chao Gou.
\newblock Learning from observer gaze: Zero-shot attention prediction oriented by human-object interaction recognition.
\newblock In {\em Proceedings of the IEEE/CVF Conference on Computer Vision and Pattern Recognition (CVPR)}, pages 28390--28400, 2024.

\bibitem{21}
Hongzhi Zhu, Septimiu Salcudean, and Robert Rohling.
\newblock Gaze-guided class activation mapping: Leverage human visual attention for network attention in chest x-rays classification.
\newblock In {\em Proceedings of the International Symposium on Visual Information Communication and Interaction (VINCI)}, pages 1--8, 2022.

\bibitem{2}
Zhengxia Zou, Keyan Chen, Zhenwei Shi, Yuhong Guo, and Jieping Ye.
\newblock Object detection in 20 years: A survey.
\newblock {\em Proceedings of the IEEE}, 111(3):257--276, 2023.

\end{thebibliography}
}

\section*{Acknowledgements}
We sincerely thank all participants who contributed to the gaze data collection process for the \textit{Gaze-CIFAR-10} dataset. In particular, we would like to acknowledge the following individuals (in no particular order): Qingzhi Zhang, Jiaxin Liu, Yue Qin, Caiyun Lu, Wenya Liu, Yanzhao Xue, Shibo Xue, Yun Xiao, Yuhua Zhou, Rui Zou, Lei Ye, Run Tongtian, Shuang Zhu, Lingyun Liu, Shuheng Li, Ming Xu, Xinyue Zhang, Haoxuan Fang, Han Su, and Yanchao Li. Their support and participation were essential to the construction of our dataset and the success of this work.

\end{document}